%% file: main.tex
\pdfoutput=1

\documentclass[11pt]{article}

\usepackage{emnlp}

\usepackage{times}
\usepackage{latexsym}

\usepackage[T1]{fontenc}

\usepackage[utf8]{inputenc}

\usepackage{microtype}

\usepackage{inconsolata}

\usepackage{graphicx}
\usepackage{tikz}
\usepackage{tipa}
\usepackage{float}
\usepackage{caption}
\usepackage{subcaption}
\usepackage{booktabs} 
\usepackage{multirow}

\title{Counterfactually Probing Language Identity in Multilingual Models}

\author{Anirudh Srinivasan$^{\diamondsuit}$\footnotemark[1]    \quad Venkata S Govindarajan$^{\heartsuit}$\footnotemark[1] \quad Kyle Mahowald$^{\heartsuit}$  \\
$^{\diamondsuit}$Department of Computer Science \quad $^{\heartsuit}$Department of Linguistics  \\ The University of Texas at Austin\\
\texttt{\{anirudhs,venkatasg,kyle\}@utexas.edu} \\
}

\begin{document}

\maketitle
\def\thefootnote{*}\footnotetext{These authors contributed equally to this work.}\def\thefootnote{\arabic{footnote}}

\input{sections/00-abstract}

\section{Introduction}
\input{sections/01-intro}

\section{Methods}
\input{sections/03-methods}

\section{Results}
\input{sections/04-results}

\section{Conclusions}
\input{sections/06-conclusion}

\bibliography{references}
\bibliographystyle{acl_natbib}

\appendix

\input{sections/99-appendix}

\end{document}

%% file: sections/00-abstract.tex
\begin{abstract}
Techniques in causal analysis of language models illuminate how linguistic information is organized in LLMs.
We use one such technique, AlterRep, a method of counterfactual probing, to explore the internal structure of multilingual models (mBERT and XLM-R).
We train a linear classifier on a binary language identity task, to classify tokens between Language X and Language Y.
Applying a counterfactual probing procedure, we use the classifier weights to project the embeddings into the null space and push the resulting embeddings either in the direction of Language X or Language Y.
Then we evaluate on a masked language modeling task.
We find that, given a template in Language X, pushing towards Language Y systematically increases the probability of Language Y words, above and beyond a third-party control language. 
But it does not specifically push the model towards translation-equivalent words in Language Y.
Pushing towards Language X (the same direction as the template) has a minimal effect, but somewhat degrades these models.
Overall, we take these results as further evidence of the rich structure of massive multilingual language models, which include both a language-specific and language-general component.
And we show that counterfactual probing can be fruitfully applied to multilingual models.
\end{abstract}

%% file: sections/01-intro.tex
Large pretrained multilingual transformer models succeed at a variety of multilingual and monolingual tasks and can be used in transfer learning paradigms, where a model is trained to do a task in one language and then transferred to another language  \citep{lauscher-etal-2020-zero,conneau-etal-2020-emerging, wu-dredze-2019-beto, wu-dredze-2020-languages, pires-etal-2019-multilingual,vulic-etal-2020-probing,rust-etal-2021-good}. 
These abilities have spurred a spate of papers probing the internal workings and capabilities of multilingual models, suggesting that such models may contain language-independent, along with langauge-specific knowledge of interesting linguistic structure \citep[e.g.,][]{chi-etal-2020-finding,papadimitriou-etal-2021-deep,ravishankar-etal-2021-attention,blevins-etal-2022-analyzing,gonen-etal-2020-greek}.

\begin{figure}
    \centering
    \includegraphics[width=\columnwidth]{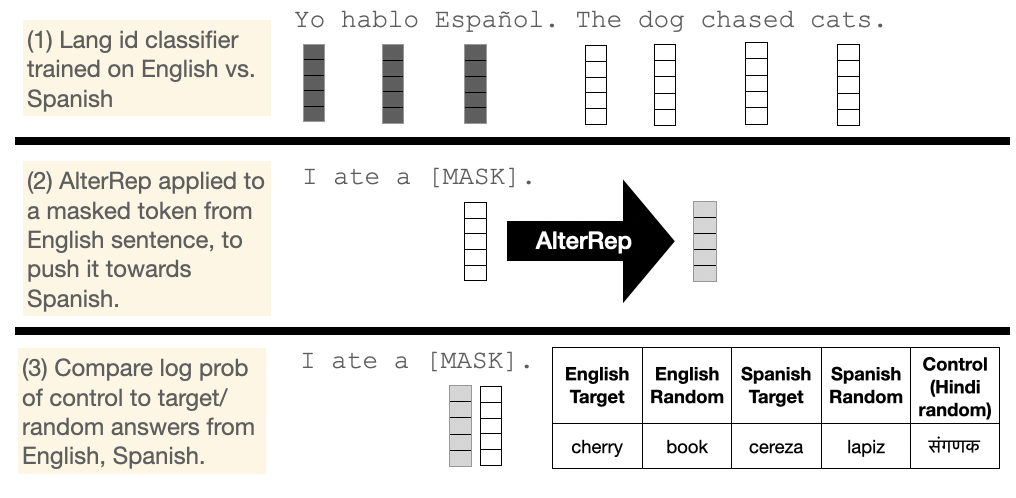}
    \caption{We train a classifier on the language ID task, and then apply AlterRep to the embeddings and examine the change in probabilities. Above, an English template sentence is pushed towards Spanish. We compare the probabilities of the target English answer to its Spanish translation-equivalent, random English and Spanish answers, and a random third-language control.}
    \label{fig:fig1}
\end{figure}

While the results of this literature are suggestive, probing methods are susceptible to memorizing the original input and may not reflect what information models actually use downstream \citep{hewitt-liang-2019-designing,elazar-etal-2021-amnesic,pimentel-etal-2020-information,voita-etal-2021-analyzing}.
It is thus desirable to test not only what information can be extracted but what information is actually used \citep{geiger2021causal,finlayson-etal-2021-causal,lasri-etal-2022-probing}.

To do that we apply AlterRep \citep{ravfogel-etal-2021-counterfactual}, an offshoot of Iterative Nullspace Projection \citep[INLP;][]{ravfogel-etal-2020-null,elazar-etal-2021-amnesic}, in a multilingual setting.\footnote{Since running these experiments, there is now work showing that linearly removing information as in INLP is sub-optimal \citep{ravfogel-etal-2022-adversarial}. A natural extension would be to explore our paradigm using these newer techniques.}
The AlterRep method is to train a classifier on the model representations to pick out a particular feature and then use the parameters learned by the classifier to \textit{intervene} on the embeddings, pushing them in a particular direction. \citet{ravfogel-etal-2021-counterfactual} use it to intervene on whether a noun phrase is in a relative clause (e.g., training a classifier on whether the noun phrase is in a relative clause and then using projections from the classifier to push the embeddings towards or away from the relative clause direction).
Crucially, they then measure how this manipulation affects downstream subject-verb number agreement.

Whereas \citet{ravfogel-etal-2021-counterfactual} use AlterRep to explore syntactic representations in models, our hypothesis is that the same kind of causal manipulation could be informative as to how multilingual models process multilingual text.
Doing so necessarily involves separating multilingual embedding space into language-neutral and language-specific components.
\citet{libovicky-etal-2020-language} explore the idea of obtaining a language-neutral representation from a multilingual model by computing an ``average'' representation for each language and subtracting it from the token embedding.

There is some precedent for using INLP to generate language-specific and language-neutral components. 
\citet{gonen-etal-2020-greek} showed that multilingual models like mBERT have both a language-specific and language-general component and that, by separating them using INLP on a language identification task, one can obtain language-agnostic representations (and, inversely, highly language-specific representations).
They show that, by training on an English vs. non-English task and then projecting onto the nullspace using INLP, the generated text on a masked language modeling task (in English) is less likely to be English after INLP.
\citet{gonen-etal-2020-greek} also show that, by subtracting an ``average'' representation of language $X$ from a particular token embedding and then adding the average language $Y$ embedding, one can obtain a translation of the token in language $Y$ by \textit{analogy}.
But they do not specifically use INLP to do these translations in a language-to-language way, as we do here.

Using a similar logic but the AlterRep technique instead of the analogical method, we test whether we can do a kind of ``translation via AlterRep'', effectively ``pushing'' the embeddings towards a particular language. First, we use the original multilingual model embeddings for a particular token $h_t$ to train a language identity classifier $C$ to classify the language of tokens from Languages $X$ and $Y$.
We then use INLP to null out language ID information, creating null embeddings $h_t^N$.
We can then generate altered embeddings $h_t^X$ and $h_t^Y$, which go beyond merely nulling out language ID and instead represent embeddings that have been pushed into the direction of Language $X$ or $Y$, respectively.
We use these counterfactual embeddings to generate predictions for masked text and compare the result to the original embeddings.

To make this concrete, imagine training a language identification classifier on English vs. Spanish, as shown in Figure~\ref{fig:fig1}.
Whereas a multilingual model would typically fill in the [MASK] position in the English sentence ``I ate a [MASK]'' with an English token, if we use the classifier to push the embeddings in the direction of Spanish, then we might expect a completion like ``I ate a \textit{cereza}'' to become more likely where \textit{cereza} is the English word for cherry.
We would expect the probability of the English word ``cherry'' to decrease.

Through this work, our hope is not only to illuminate the innerworkings of multilingual models, but also to validate and explore the use of counterfactual probing in a novel domain.

To spoil the result: we show that language identity is encoded in contextual token embeddings and, crucially, that this information is \textit{used} by multilingual models in masked language modeling.
In effect, pushing embeddings in the direction of a particular language (and away from another) systematically increases probabilities of words in the \textsc{PushedTo} language and decreases the probabilities in the \textsc{PushedAway} language, while leaving words from other languages unchanged.
By comparing the changes in probabilities of target words in the \textsc{PushedTo} language (i.e., translation equivalents of the original correct word) to random words in that language, we see that our alterations seem to push the model towards the \textit{prior} of the intended language, without specifically boosting the semantic equivalent.\footnote{We make our code available online \href{https://github.com/venkatasg/multilingual-counterfactual-probing}{\texttt{here}}.}

%% file: sections/03-methods.tex
We run two experiments, with slightly different procedures. In Experiment 1, we train a token-level language ID classifier on a corpus of monolingual sentences from 2 languages, without mixing the languages within-sentence. In Experiment 2, we create artificial code-mixed text (mixing within sentences) and use this for training the classifier. In both experiments, we evaluate two representative massive multilingual transformer models, Multilingual BERT \cite[mBERT;][]{devlin-etal-2019-bert} and XLM-RoBERTa Base \cite[XLM-R;][]{conneau-etal-2020-unsupervised}, and we focus on the last layer for intervention.
We describe each step in more detail below.

\paragraph{Models}  Multilingual BERT \cite{devlin-etal-2019-bert} and XLM-Roberta Base \cite{conneau-etal-2020-unsupervised} span 104 and 100 languages respectively. Both are transformer encoders that have a hidden dimension size of 768.

\paragraph{Classifier} For each iteration of INLP, a linear classifier is learned on the representations produced by the encoder to predict language ID ($L_1$ vs $L_2$) for each token in the input. We use SVMs as our linear classifiers \citep[as in][]{ravfogel-etal-2021-counterfactual}. While training the classifier, 15\% of the tokens are randomly masked. This is done to be more representative of the final evaluation setting where masked inputs are used. The classifier is trained on balanced samples.

\paragraph{INLP and AlterRep}
INLP is a technique for removing information from embeddings.
Specifically, INLP uses the weights learned by each classifier to project the embedding $h_t$ onto the intersection of nullspaces of the classifiers $h_t^N$ (this contains no information for doing the classification). The component orthogonal to this $h_t^R$, contains all of the information for doing classification. 
In practice, not all information is removed by the first projection onto the nullspace, so the process is repeated iteratively.
The second classifier is learned on top of the embeddings whose information has been nulled out based on the first classifier's weights, and so on. This is repeated $m$ times, yielding $m$ classifiers.

AlterRep \cite{ravfogel-etal-2021-counterfactual} considers both the nullspace component and the orthogonal component to generate a new embedding $h_t'$ that has been modified to lie on a particular side of the classifier. Suppose that for weight $w_i$ learned by classifier $i$, $h_t^{w_i}$ is the orthogonal component.
The counterfactual vector $h_t'$ is created as follows:

\begin{equation}
    h_t' = h_t^N + \alpha \sum_{w_i} S * h_t^{w_i}
\end{equation}

\noindent $S$ is 1 when the given classifier's prediction $w_i^T h_t>0$ (predicts $L_1$) and -1 when $w_i^T h_t<0$ (predicts $L_2$).

The parameter $\alpha$ controls the direction and magnitude of the alteration. 
When $\alpha = 0$, it's equivalent to amnesic probing. While training classifiers for INLP, $\alpha$ is always set to 0. $\alpha$ is non zero when we're evaluating on MLM in the subsequent sections.
When $\alpha > 0$, the representations will be pushed to the $L_1$ side of the classifier, irrespective of where they were originally.
When $\alpha < 0$, the representations will be pushed to the $L_2$ side of the classifier, irrespective of where they were originally.

\paragraph{Choosing the number of INLP iterations} Determining the number of iterations to run INLP for is tricky as there is tension between removing information and destroying the language model \citep{elazar-etal-2021-amnesic}.
We sought to find a number of iterations that would (a) significantly degrade performance on the language identification task (thus proving removal of language ID information) but (b) not torpedo the performance of the model on the MLM task.

\begin{figure*}
    \centering
    \includegraphics[width=\textwidth]{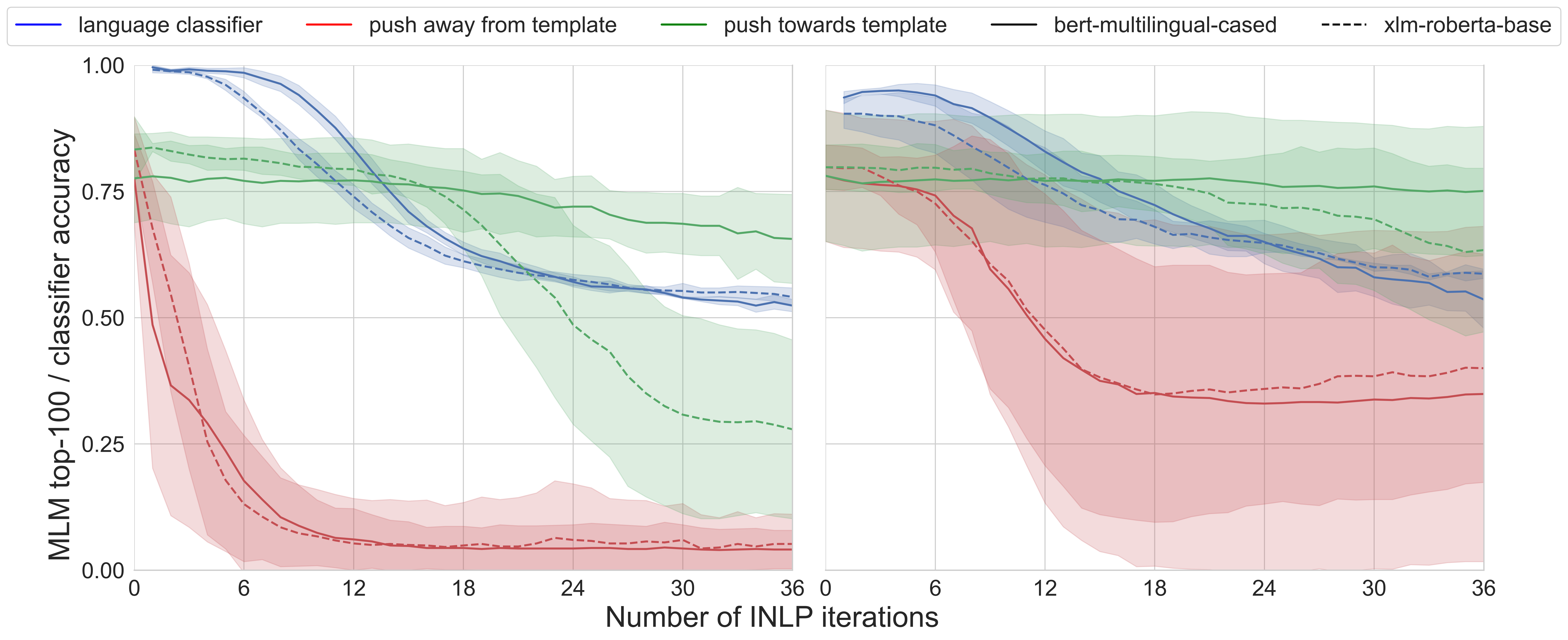}
    \caption{MLM-100 accuracies after intervention, and language ID classifier accuracy plotted over number of INLP iterations for m-BERT and XLM-R. Results are shown with INLP trained on non-code mixed data on the left, and code-mixed data on the right. All MLM results are accuracies averaged over all languages and language pairs}
    \label{fig:lmacc-iters}
\end{figure*}

One option for choosing the number of iterations to run INLP is to run it until the classifier performance is at chance on the target task.
We found that, if we do this for XLM-R (and to a lesser extent for mBERT), a large number of iterations is required (around 32).
This large number of iterations effectively destroys the language model, causing the most likely completions to be jibberish (with a MLM-100 accuracy close to zero).

So, instead, we choose to optimize for removing as much information as possible while still maintaining acceptable (>90\%) MLM-100 accuracy.
Figure \ref{fig:lmacc-iters} shows the number of iterations plotted against both the MLM-100 measure and against the language ID accuracy.
For Experiment 1, we chose 4 iterations for XLM-R and mBERT.
For Experiment 2, we run for more iterations (16 for both models) since the code-mixed data is less susceptible to model degradation.

Note that this means that, for our post-INLP models, there is still some language identity information remaining and so these embeddings should not be treated as entirely free of language identity information.
But the number of iterations was still sufficiently high to allow us to meaningfully push towards or away from the original language.

Running the INLP classifier for the same number of iterations more catastrophically affects the overall MLM performance for XLM-R than it does mBERT.
We leave it to future work to ascertain why XLM-R might have its MLM performance more closely tied to language identity information than mBERT.

\subsection{Experiment 1: Non-Code-Mixed Sentences}

\paragraph{Languages} We pair English with each of Korean, Hindi, Spanish, and Finnish, giving us 4 pairwise comparisons. These languages were chosen to form pairs with the same script/family (English-Spanish), same script but different family (English-Finnish) and different script/family (English with Hindi and Korean). We always use English as one of the pairs, which ensures adequate translations using the MUSE dictionaries. But see Experiment 2 for results between non-English pairs.

Table \ref{tab:mono_data} shows the sources and statistics for the data used to train these classifiers. The monolingual sentences for English and Hindi are taken from their corresponding parts of an English-Hindi parallel corpus \cite{kunchukuttan-etal-2018-iit}. The data for Korean is taken from ParaCrawl \cite{espla-etal-2019-paracrawl}, Spanish and Finnish from EuroParl \cite{koehn-2005-europarl}.

\paragraph{Training/Testing methodology} The Language ID classifiers are trained using 1500 sentences from each language.
We alternately embed sentences from English and sentences from the other language and then extract the token embeddings. The classifier learns to predict whether a given token is extracted from the English or non-English language.

\begin{table}[t]
\centering
\begin{tabular}{@{}llllll@{}}
\toprule
                       & Lang & Source     & Train & Val & Test \\ \midrule
\multirow{2}{*}{En-Es} & En   & IITB En-Hi & 1500  & 250 & 250  \\
                       & Es   & EuroParl   & 1500  & 250 & 250  \\ \midrule
\multirow{2}{*}{En-Fi} & En   & IITB En-Hi & 1500  & 250 & 250  \\
                       & Fi   & EuroParl   & 1500  & 250 & 250  \\ \midrule
\multirow{2}{*}{En-Hi} & En   & IITB En-Hi & 1500  & 250 & 250  \\
                       & Hi   & IITB En-Hi & 1500  & 250 & 250  \\ \midrule
\multirow{2}{*}{En-Ko} & En   & IITB En-Hi & 1500  & 250 & 250  \\
                       & Ko   & ParaCrawl  & 1500  & 250 & 250  \\ \bottomrule
\end{tabular}
\caption{Monolingual Data Sources/Sizes}
\label{tab:mono_data}
\end{table}

Evaluation of AlterRep is done on a target of 250 sentences from each language, from the test sets of the same corpora used for training the language ID classifiers.
But, because we cannot always find a dictionary match for each target word, the number of test sentences ranges in practice from 205 to 243.
We take sentences from the language ID classifier test sets and randomly pick a word to mask in each sentence. We treat that word as the \textit{target word} in the original language, and we use MUSE dictionaries \citep{DBLP:conf/iclr/LampleCRDJ18} to find the equivalent of that word in the alternate language.
Then, we compare the probability of (a) the target word in the original language, (b) the target word in the other language, (c) a random word in the original language, (d) a random word in the other language, and (e) a random word form a \textit{third} language (which serves as a control). 
For instance, Figure \ref{fig:fig1} shows an English sentence ``I ate a cherry.'' where we mask the token ``cherry.''. Table \ref{tab:exp1_example} shows an example of how we modify the masked word in the sentence in different manners.
\begin{table}[t]
\small
    \centering
    \begin{tabular}{p{0.5\linewidth}  p{0.4\linewidth}}
        \toprule
        \textbf{Original sentence} & I ate a \textit{cherry} \\  \midrule
        \textbf{Masked input to model} & I ate a \texttt{[MASK]} \\ \midrule
        \textbf{Mask replaced with target language (es) word} & I ate a \textit{cereza} \\ \midrule
        \textbf{Mask replaced with random target language (es) word} & I ate a \textit{lapiz} \\ \midrule
        \textbf{Maks replaced with third language (fi) word} & I ate a \textit{kirsikka} \\ \bottomrule
    \end{tabular}
    \caption{Example of how we replace a masked word with different words from the target language/third language dictionary}
    \label{tab:exp1_example}
\end{table}

When we push that masked token in the direction of Spanish using AlterRep, we then compare the log probability (before and after the intervention) of: the target English word (``cherry''), the Spanish translation-equivalent (``cereza''), a randomly chosen English word, a randomly chosen Spanish word, and a randomly chosen control word from a third language.
The random words are all chosen to have the same number of tokens as the target word in that language.
As is standard, we obtain log probabilities for multi-token words by averaging \citep{kassner-etal-2021-multilingual,dou-neubig-2021-word}.

If the AlterRep procedure works, then if we start with an English template and push the masked token towards Spanish, the probability of Spanish words will rise and the probability of English words will decrease, while the probability of Hindi words will be unaffected.
When we start with English and push towards English, we expect little change.
If there are shared semantic representations across languages, then we might expect to see the target words in the pushed-towards language (e.g., ``cereza'', Spanish ``cherry'') increase more than random ones (e.g., ``lapiz'', Spanish for ``pencil'').

\subsection{Experiment 2: Mixed-Language Sentences}
\paragraph{Languages} To assess the robustness of our results, we focus on a scenario where the model is exposed to mixed-language text, as opposed to monolingual text. Existing work \cite{santy-etal-2021-bertologicomix} has probed the abilities of multilingual transformer encoders on code-mixed text and has shown that these models are able to learn language ID in code-mixed scenarios and this experiment serves as a further probe into the cross-lingual abilities of these models. We consider 3 languages: English, Hindi and Korean and consider all 3 pairs using these languages (En-Hi, En-Ko and Hi-Ko).

\paragraph{Training/Testing methodology} The language ID classifiers are trained using synthetic code-mixed text generated for these 3 language pairs.
Generating training data this way gives us the flexibility in evaluating on any language pair that we want (unlike using real code-mixed which would limit the language pairs we could choose). We created the synthetic code-mixed data by lexical substitution of words in a monolingual sentence using MUSE dictionaries \citep{DBLP:conf/iclr/LampleCRDJ18}, substituting so that ~30\% of the words are in the second language. Table \ref{tab:cm_data} shows the sources and the statistics for the data used to train this.

Evaluation is done using the multilingual mLAMA dataset \cite{kassner-etal-2021-multilingual}. Based on Wikipedia entity relations, it consists of templates, translated across languages, with slots in which masked language modeling has to be used to fill in the correct mLAMA answer. Thus, in this experiment, the masked token is always the mLAMA answer in a particular language instead of a random word. We thus have the same template in both languages, along with correct answers in both languages that we can use to evaluate AlterRep on. The number of templates used for evaluation are n=7,256 for English-Hindi, 14,204 for English-Korean, 6,496 for Hindi-Korean.
Because we are not limited to pairs involving English in this experiment, we focus on all pairwise comparisons between Hindi, English, and Korean for this study 

\begin{table}[t]
\begin{tabular}{@{}llllll@{}}
\toprule
                       & Lang & Source                                                      & Train                 & Val                  & Test                 \\ \midrule
\multirow{2}{*}{En-Hi} & En   & IIT En-Hi                                                   & \multirow{2}{*}{3000} & \multirow{2}{*}{500} & \multirow{2}{*}{500} \\
                       & Hi   & \begin{tabular}[c]{@{}l@{}}Word Subn\\ w/ MUSE\end{tabular} &                       &                      &                      \\ \midrule
\multirow{2}{*}{En-Ko} & En   & IIT En-Hi                                                   & \multirow{2}{*}{3000} & \multirow{2}{*}{500} & \multirow{2}{*}{500} \\
                       & Ko   & \begin{tabular}[c]{@{}l@{}}Word Subn\\ w/ MUSE\end{tabular} &                       &                      &                      \\ \midrule
\multirow{2}{*}{Hi-Ko} & Hi   & IIT En-Hi                                                   & \multirow{2}{*}{3000} & \multirow{2}{*}{500} & \multirow{2}{*}{500} \\
                       & Ko   & \begin{tabular}[c]{@{}l@{}}Word Subn\\ w/ MUSE\end{tabular} &                       &                      &                      \\ \bottomrule
\end{tabular}
\caption{Code Mixed Data Sources/Sizes. To generate code-mixed data, text from the first language is taken and words from the second language using the MUSE dictionary}
\label{tab:cm_data}
\end{table}

%% file: sections/04-results.tex
\begin{figure*}[t]
    \begin{subfigure}[b]{1\columnwidth}    
        \centering
        \includegraphics[width=1\columnwidth, trim={0 22 0 0}, clip]{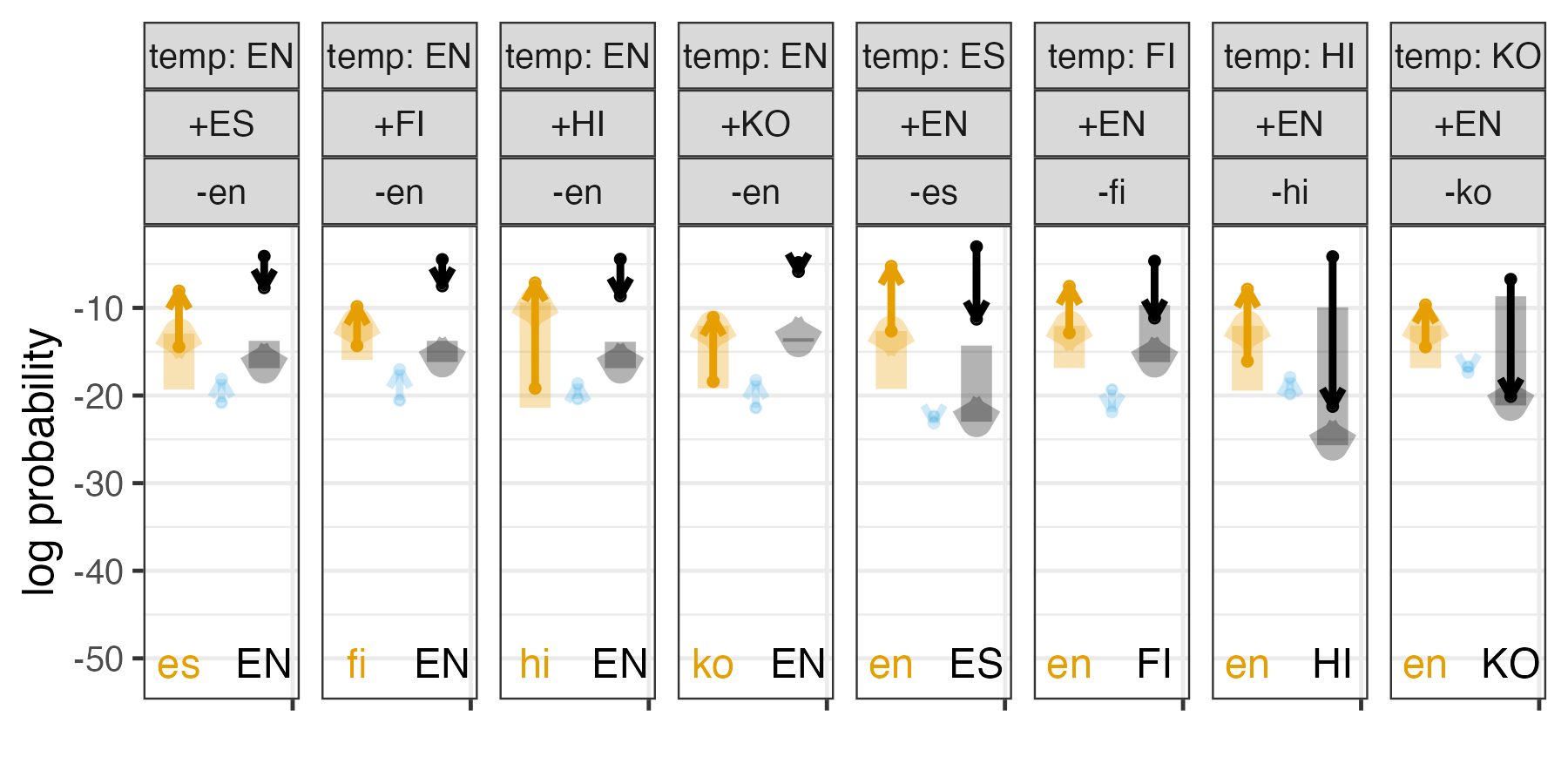}
        \label{fig:overall}   
        \vspace{-1em}
        \caption{mBERT, pushing tokens in opp direction to template}
    \end{subfigure}
    \begin{subfigure}[b]{1\columnwidth}    
         \centering
        \includegraphics[width=1\columnwidth, trim={0 22 0 0}, clip]{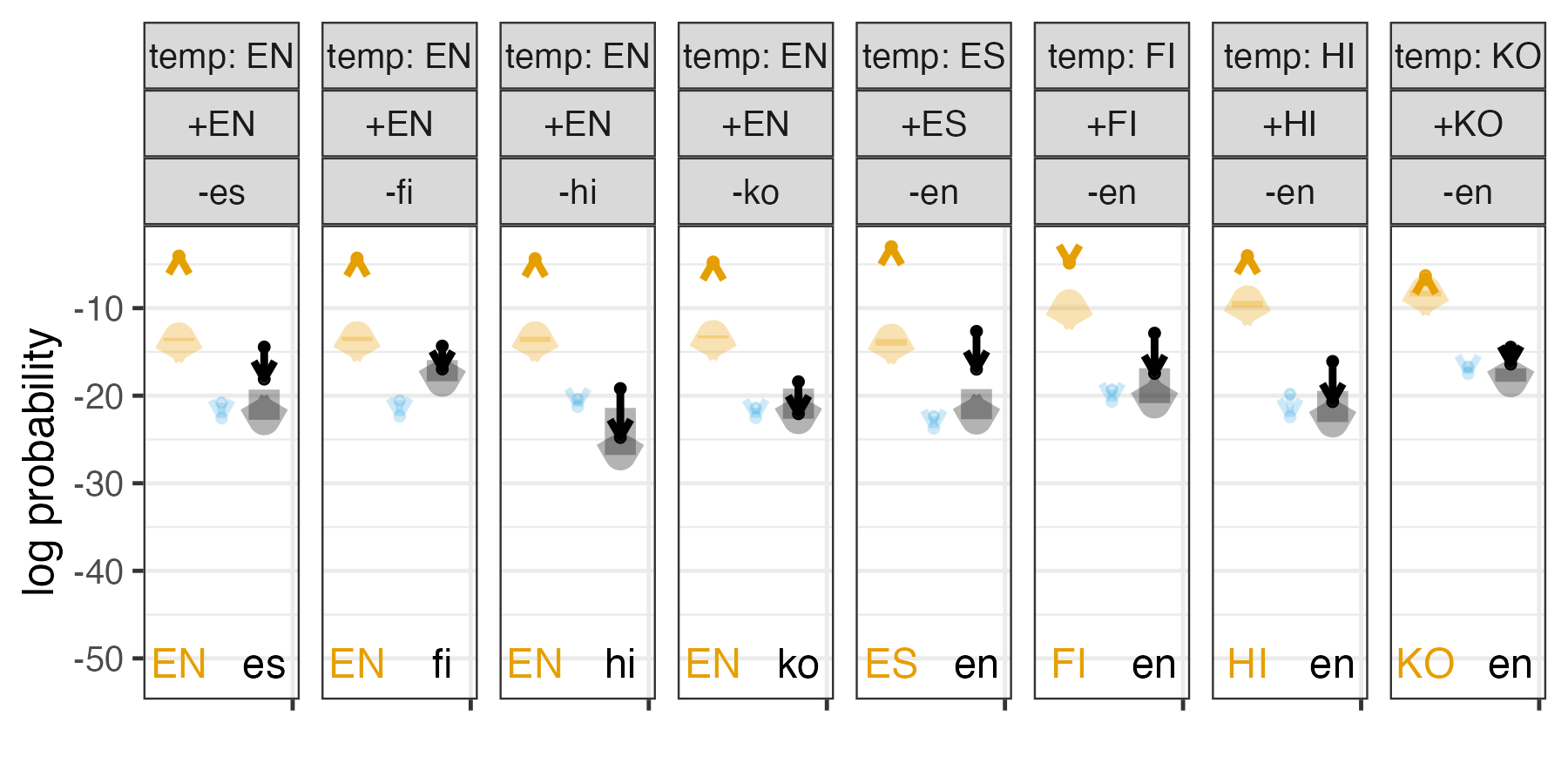}
        \label{fig:agree}
        \vspace{-1em}
        \caption{mBERT, pushing tokens in same direction as template}
    \end{subfigure}
    \\
    \begin{subfigure}[b]{1\columnwidth}
        \centering
        \includegraphics[width=1\columnwidth, trim={0 22 0 0}, clip]{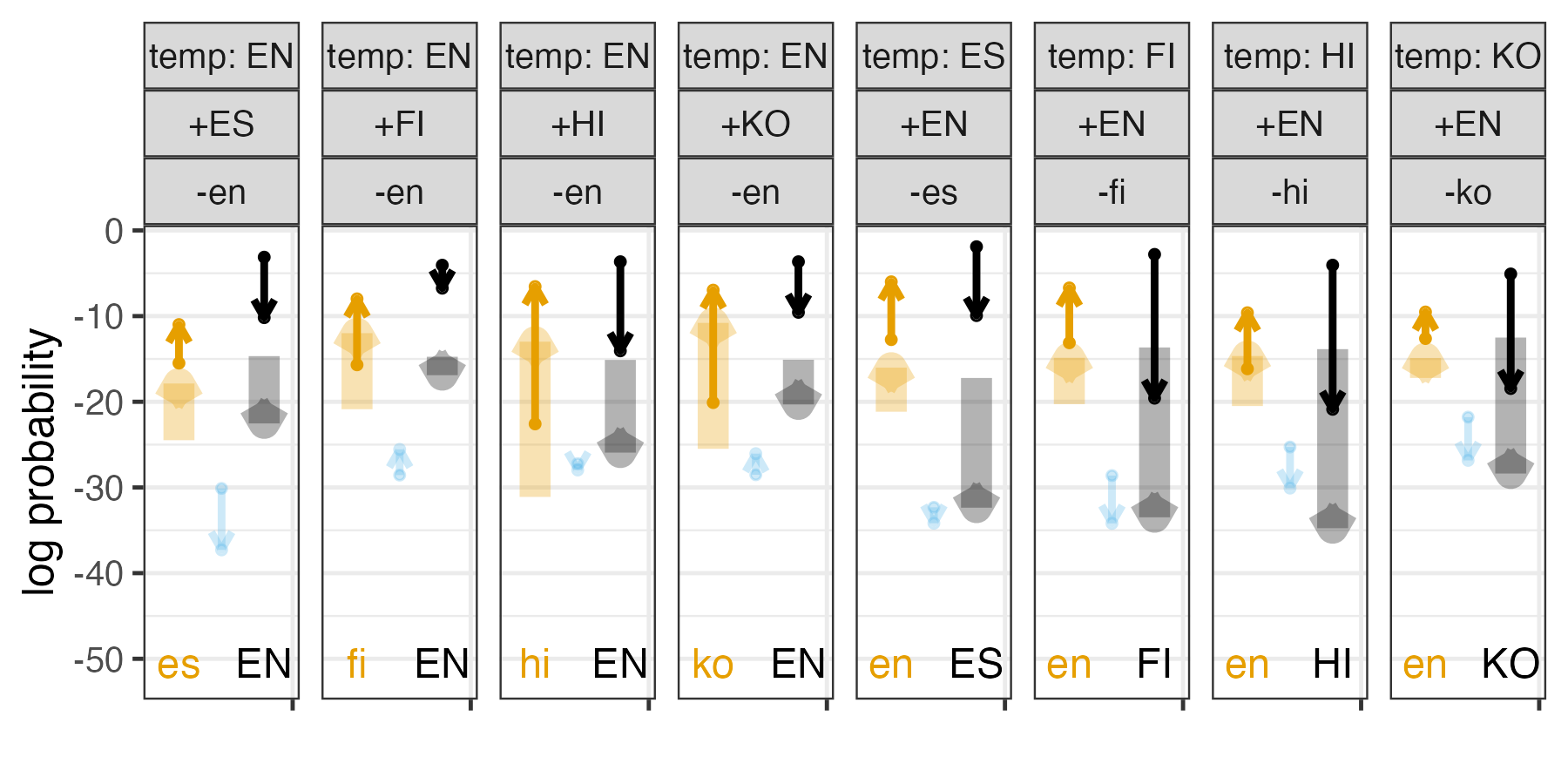}
        \label{fig:overall-3}
        \vspace{-1em}
        \caption{XLMR, pushing tokens in opp direction to template}
    \end{subfigure}
    \begin{subfigure}[b]{1\columnwidth}
         \centering
        \includegraphics[width=1\columnwidth, trim={0 22 0 0}, clip]{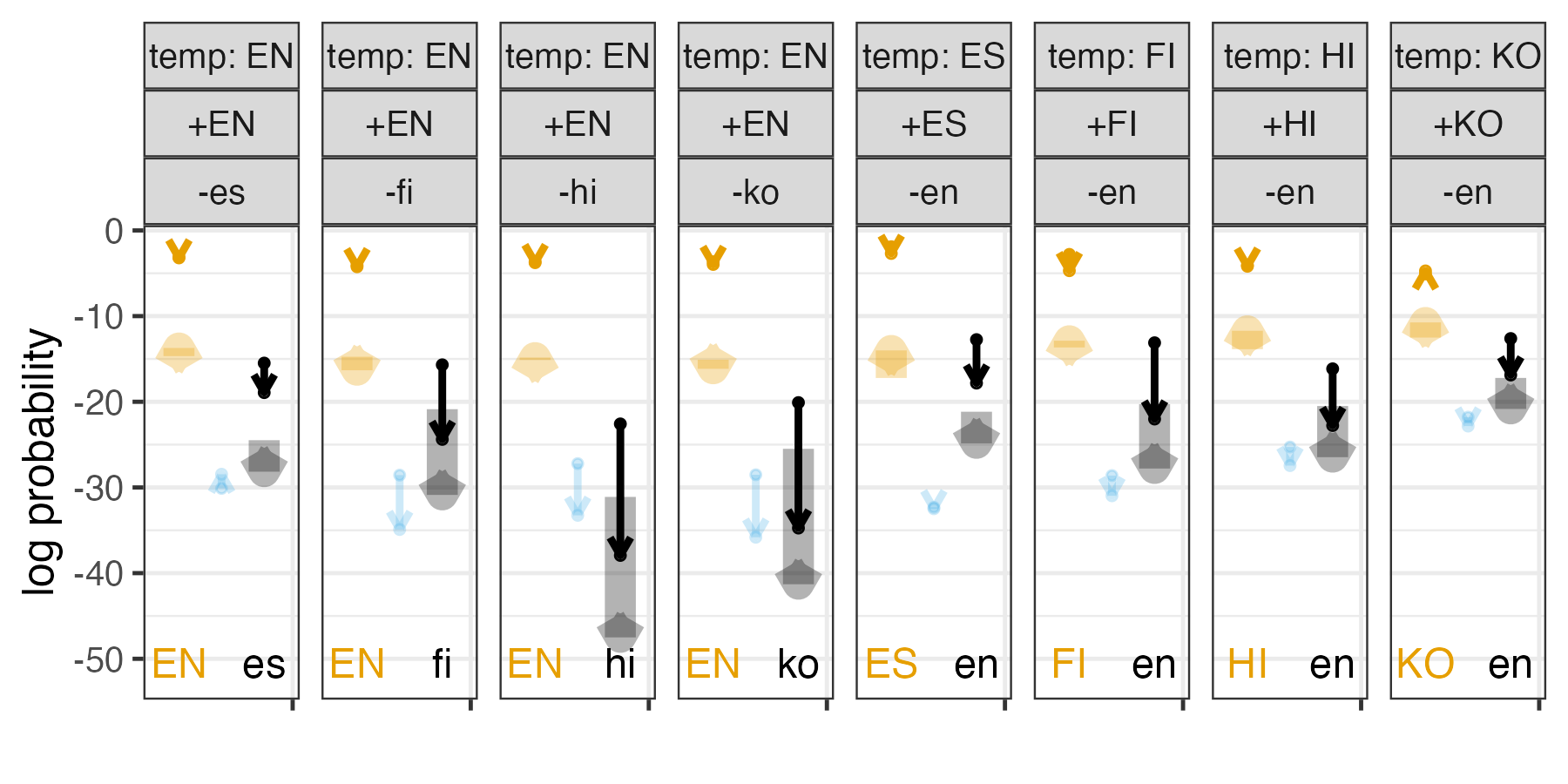}
        \label{fig:agree-5}
        \vspace{-1em}
        \caption{XLMR, pushing tokens in same direction as template}
    \end{subfigure}
    \vspace{-0.5em}
    \caption{Change in language-specific probability distributions for Exp. 1. When we push the token in the opposite language of the template (left two figures), we can see significant changes in the probability distributions for the target (dark arrows) and random words (shaded arrows) from that language, with some cases showing such a large change that tokens from the new language have more probability and will be sampled. Third language controls (blue arrows) and pushing tokens in the same language as the template (right 2 figures) don't show much change.}
    \label{fig:mainresults}
\end{figure*}

\begin{table}[ht]
\small
 \centering
 \begin{tabular}{p{1cm}p{1cm}p{1cm}p{1cm}p{1cm}}
   \hline
  \textbf{Push in dir. of temp.} & \textbf{Answer pushed towards} & \textbf{Third Lang} & \textbf{Target Word} & \textbf{Random Word} \\
   \hline
         \multicolumn{5}{c}{\textbf{mBERT}} \\
    \hline
    Opposite & Opposite  & 0.66 & 0.98 & 0.93 \\
     Opposite & Same & - & 0.98 & 0.98 \\
    Same & Opposite  & 0.10 & 1.00 & 0.99 \\
     Same & Same & - & 0.54 & 0.77 \\
   \hline
      \multicolumn{5}{c}{\textbf{XLMR}} \\
         \hline

    Opposite  & Opposite  & 0.37 & 0.99 & 0.96 \\
    Opposite  & Same & - & 0.92 & 0.92 \\
    Same & Opposite  & 0.25 & 1.00 & 0.98 \\
    Same & Same & - & 0.36 & 0.62 \\
    \hline
 \end{tabular}
 \caption{Exp 1. Proportion of data points that move in the expected direction, as a function of template matching push direction and answer matching push direction. When ``push in dir. of temp'' says ``opposite'', that means we are pushing away from the direction of the template (e.g., pushing an English sentence to Hindi).
When ``push in dir. of temp says ``same'', that means we are pushing in the same direction of the template (e.g., pushing an English sentence even further toward English).
We break down how often an answer word moves in the expected direction when that answer word is being pushed towards (e.g., an English word in a template that is being pushed towards English) or when that answer word is being pushed away from (e.g., an English word in a template that is being pushed toward Hindi). The Target word is the actual template word or its translation-equivalent. The random word is a random word in the same language. The third-party word is a random word in a third-party language.}\label{tab:exp1}
 \end{table}

\input{tables/example}

\begin{table}[ht]
\small
 \centering
\begin{tabular}{p{1.5cm}p{1.5cm}p{1cm}p{1cm}}
   \hline
  \textbf{Push in dir. of temp.} & \textbf{Answer pushed towards} & \textbf{Target Word} & \textbf{Random Word} \\
   \hline
   \multicolumn{4}{c}{\textbf{mBERT}} \\
      \hline

       Opposite & Opposite  & .90 & .87 \\
     Opposite & Same & 0.98 & 0.98  \\
    Same & Opposite  & 1.00 & 1.00 \\
     Same & Same & 0.46 & 0.64 \\
   \hline

   \multicolumn{4}{c}{\textbf{XLMR}} \\
   \hline
    Opposite & Opposite  & .99 & .96 \\
     Opposite & Same & 0.95 & 0.86  \\
    Same & Opposite  & 1.00 & 1.00 \\
     Same & Same & 0.41 & 0.37 \\
    \hline
 \end{tabular}

 \caption{Proportion of data points that move in the expected direction, as a function of the template matching push direction and answer matching push direction. When ``push in dir. of temp'' says ``opposite'', that means we are pushing away from the direction of the template (e.g., pushing an English sentence to Hindi).
When ``push in dir. of temp says ``same'', that means we are pushing in the same direction of the template (e.g., pushing an English sentence towards English).
We break down how often an answer word moves in the expected direction when that answer word is being pushed towards (e.g., an English word in a template pushed towards English) or when that answer word is being pushed away from (e.g., an English word in a template that is being pushed toward Hindi). The Target word is the actual template word or its translation-equivalent. The random word is a random word in the same language.}\label{tab:exp2}
\end{table}

Overall, across both Experiments, we find that the AlterRep operation works as expected in the majority of cases.
Figure~\ref{fig:mainresults} shows data for our Experiment 1, on mBERT 
and {XLM-~R}. In each subfigure, the top row indicates the language of the template, the 2nd row indicates the direction in which the token embedding is pushed. The plot has dark arrows indicating the change in probability distributions of tokens from the 2 languages (as indicated), with shaded arrows indicating changes for random tokens in those languages. Blue arrows indicate change in probability distributions for random tokens.

We consider separately the case where we push in the opposite direction as the template (e.g., pushing a Korean template in the English direction)  (the left 2 subfigures indicate this) vs. the case where we push in the same direction (the right 2 subfigures indicate this).
In the analysis below, we focus on the proportion of time that the probabilities shift in the expected direction after the intervention.
The mean change in log probability, before and after intervention, tells a similar story and is shown in Figure \ref{fig:mainresults}. 

From hereon, we focus on $\alpha=3$, but see Appendix~\ref{app:alpha} for results on sensitivity to this parameter.

\subsection{Experiment 1}
\textbf{When we push in the \textit{opposite direction} of the template} (e.g., push an English template towards Spanish), \textbf{the template language probabilities plummet}, both for the target (99\% of the time, across pairs) and random words (93\% of the time, across pairs). 
The fact that the target word decreases more than the random one may not be very meaningful: the target word starts out with very high probability and so it has farther to drop.
Crucially, \textbf{the \textsc{PushedTo} language probabilities all increase significantly} (98\% of the time for target answers, 98\% of the time for random answers).
\textbf{The \textsc{ThirdLang} control words show little change}, as predicted (decreasing 66\% of the time).
Thus, this manipulation works as expected: taking a mask from an English language template and pushing it towards Spanish causes the probability of all Spanish words to increase while decreasing the probability of English-language words and leaving other language words (e.g., Korean or Hindi) largely untouched.

\textbf{When we push in the \textit{same direction} as the template} (e.g., we push an English template even further in the English direction), we find that the \textbf{\textsc{OriginalLanguage} is largely unchanged} (increasing in 54\% of pairs for target words and 77\% of the time for random words).
Here the difference between random and target is likely because the target word is already at ceiling.
\textbf{The \textsc{PushedAway} language drops significantly} for both target and random words (decreases for 100\% and 99\% of pairs, for both random and target words). 
\textbf{The \textsc{ThirdLang} control decreases} 90\% of the time, suggesting that the probability of a third party language becomes even less likely when we push in the same direction as the template.
Taken together, these results suggest that pushing in the same direction as the template does not make the language model better (the target word does not increase substantially), but it does make it more likely to generate words from that language.
That is, if we push towards English and the target answer is ``dog'', pushing towards English will not make ``dog'' more likely but it will increase the overall Englishness in the model, essentially pushing it towards the English prior while decreasing the probability of generations in other languages.

Table~\ref{tab:exp1} summarizes these results, showing the fraction of templates for which the probabilities move in the expected direction.
We see movement in the expected direction in all cases except on words in the pushed-towards language, when we push in the direction of the template. That is, English words don't become \textit{even more likely} when we push towards English in an English template.
These results are consistent, regardless of whether we have a language pair with the same script (e.g., English and Finnish) or pairs with different scripts (e.g., English and Hindi).
Given the large overlap in shared tokens between any two Latin script languages (and low overlap across scripts), this consistency is notable.

\begin{figure*}[t]
    \begin{subfigure}[b]{1\columnwidth}
        \centering
        \includegraphics[width=1\columnwidth, trim={0 22 0 0}, clip]{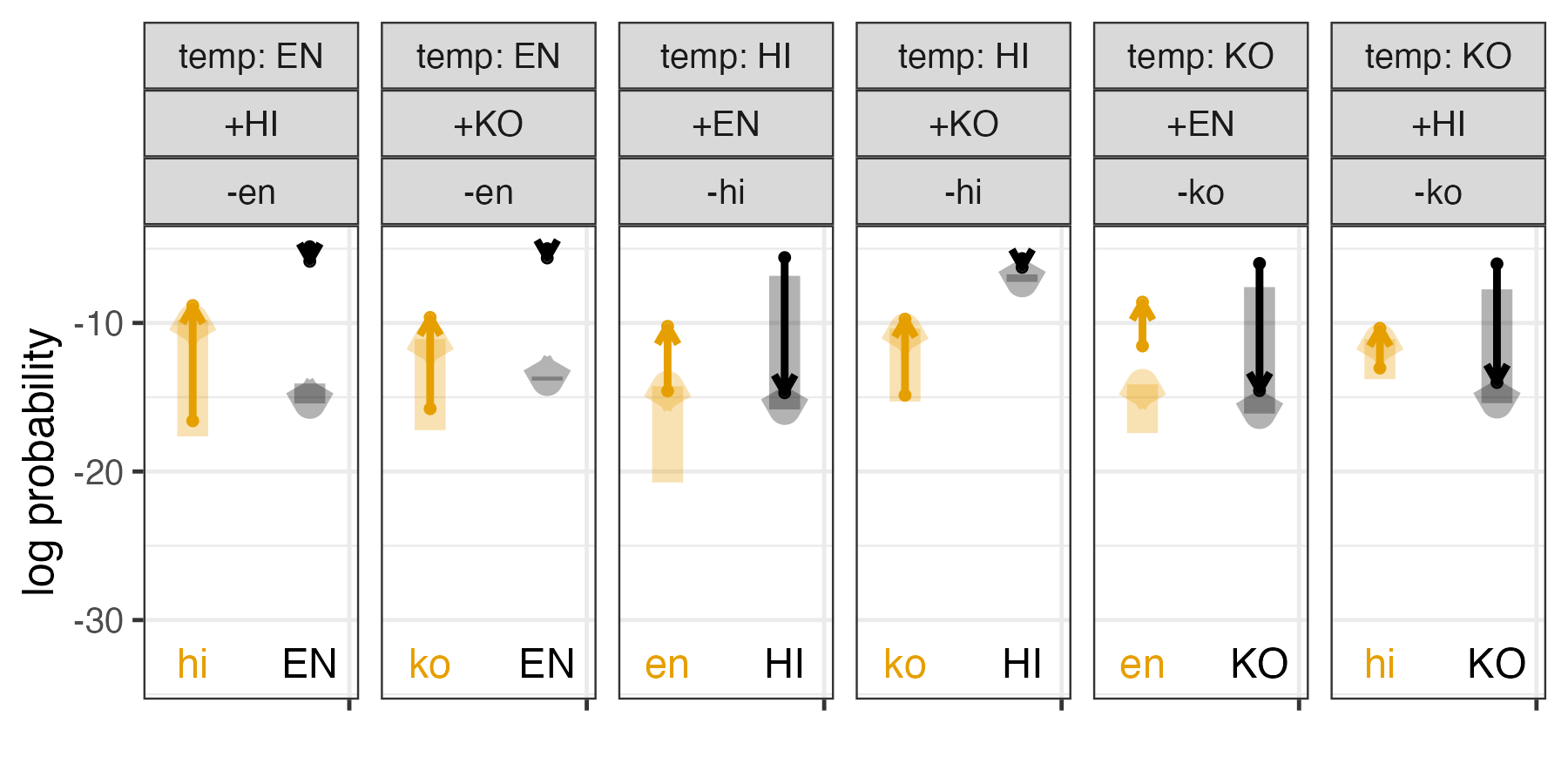}
        \label{fig:overall-2}
        \vspace{-1em}
        \caption{mBERT, pushing tokens in opp direction to template}
    \end{subfigure}
    \begin{subfigure}[b]{1\columnwidth}
        \centering
        \includegraphics[width=1\columnwidth, trim={0 22 0 0}, clip]{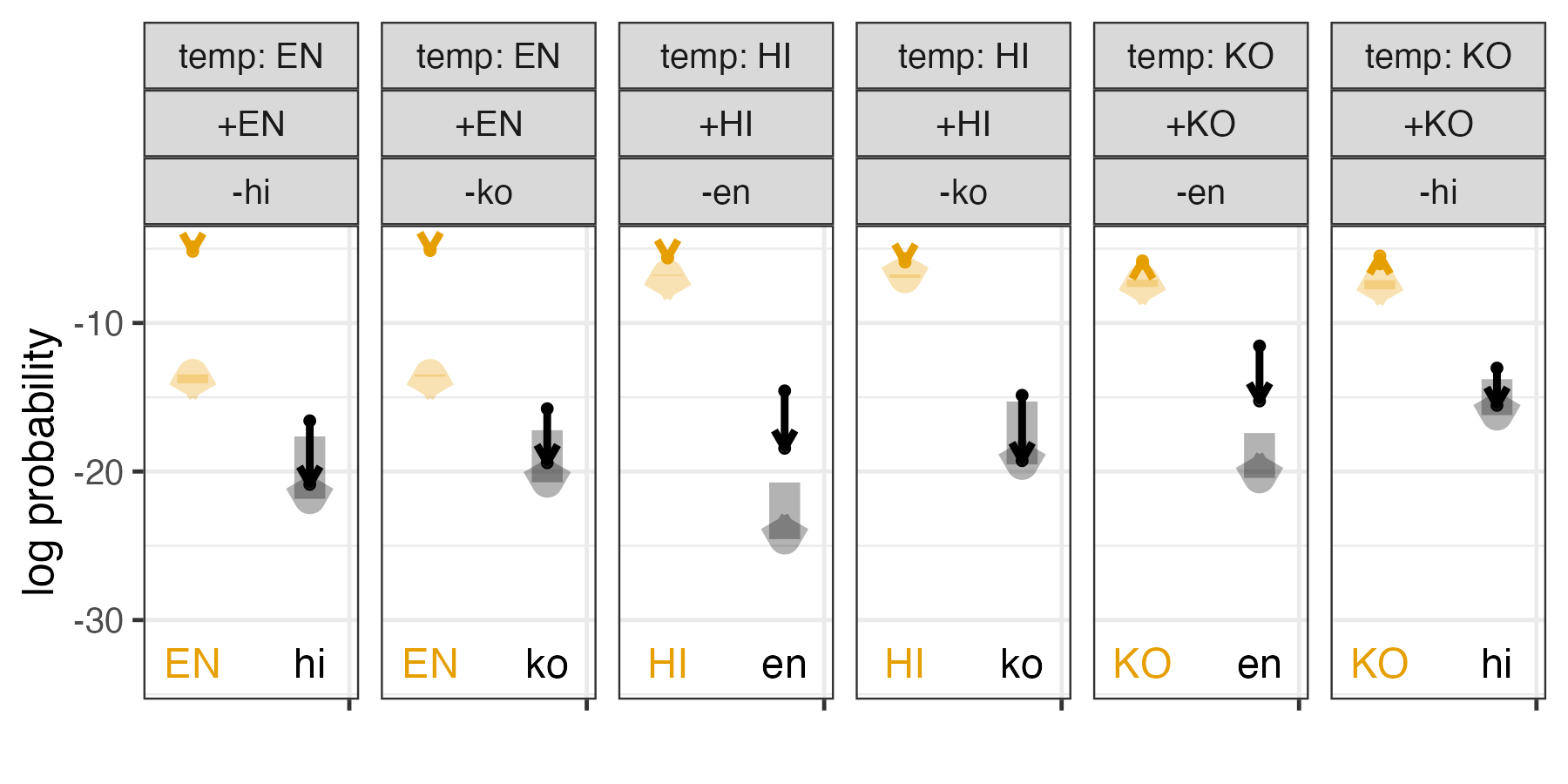}
        \label{fig:agree-2}
        \vspace{-1em}
        \caption{mBERT, pushing tokens in same direction as template}
    \end{subfigure}
    \\
    \begin{subfigure}[b]{1\columnwidth}
        \centering
        \includegraphics[width=1\columnwidth, trim={0 22 0 0}, clip]{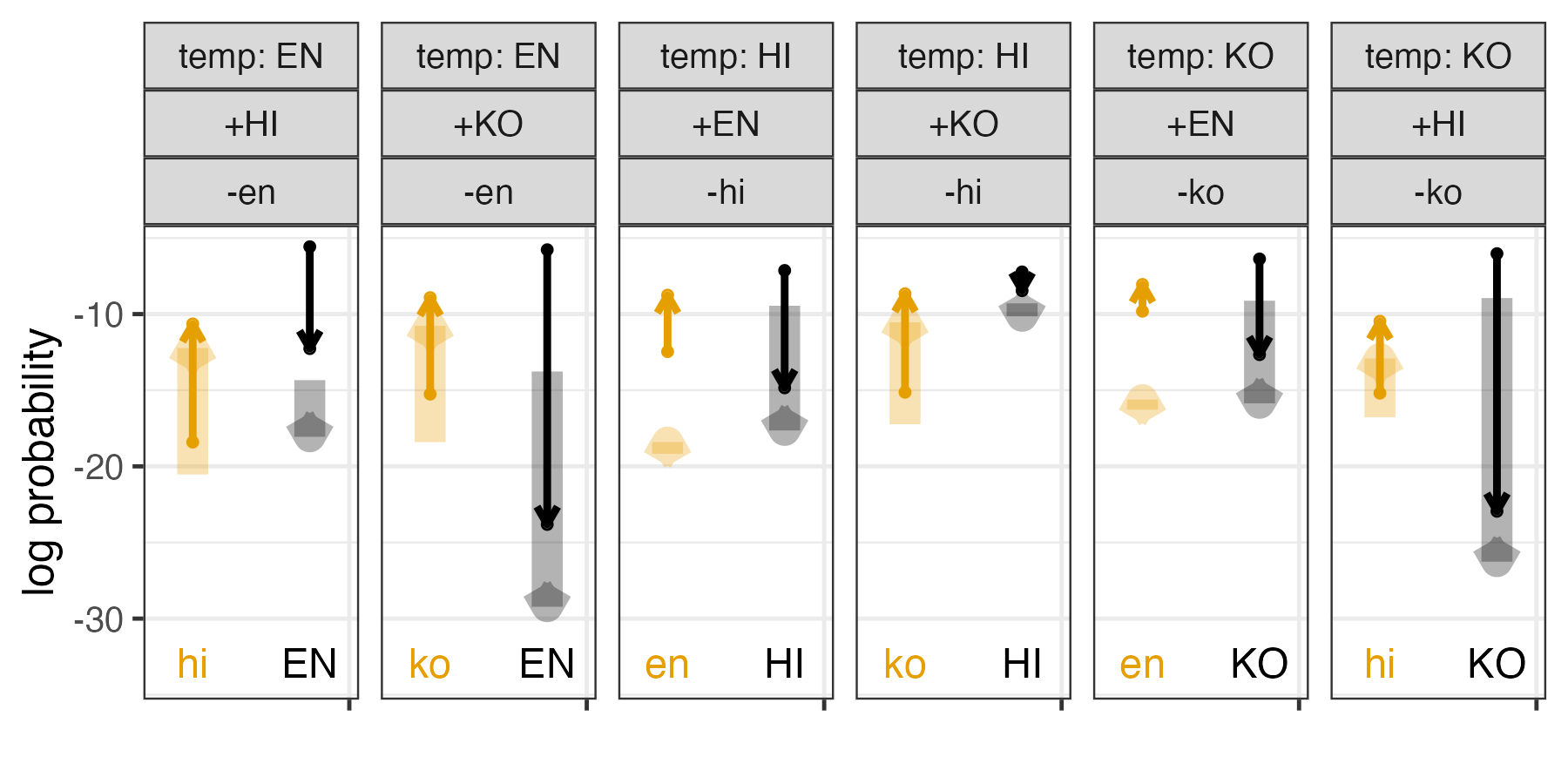}
        \label{fig:agree-3}
        \vspace{-1em}
        \caption{XLMR, pushing tokens in opp direction to template}
    \end{subfigure}
    \begin{subfigure}[b]{1\columnwidth}
        \centering
        \includegraphics[width=1\columnwidth, trim={0 22 0 0}, clip]{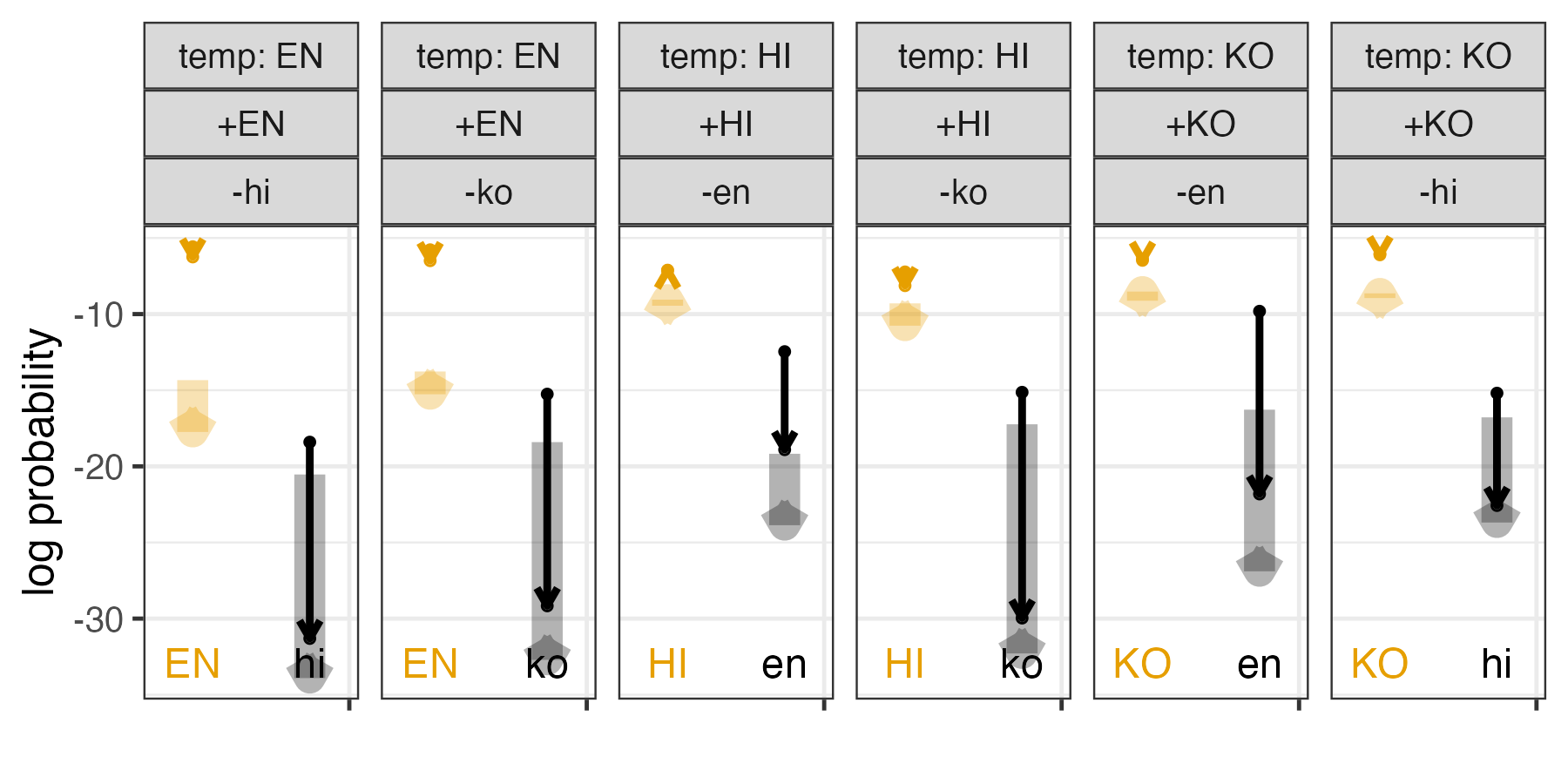}
        \label{fig:agree-4}
        \vspace{-1em}
        \caption{XLMR, pushing tokens in same direction as template}
    \end{subfigure}

    \vspace{-0.5em}
        \caption{Change in language-specific probability distributions for Exp. 2. As with Exp. 1, when we push the tokens in the opposite direction to the template (left two plots), there are bigger changes in the probability distribution, with the new language sometimes having higher probabilities than the original one. Pushing in the same direction as the template (right two plots) doesn't show any change in the ordering of the two languages.}
        \label{fig:exp2plot}
\end{figure*}

\subsection{Experiment 2}

Experiment 2 is notable for the fact that we're evaluating the model in a code-mixed setting and testing the model on queries that involve real world factual knowledge (the relations in mLAMA).
Results are similar for Experiment 2 (see Figure~\ref{fig:exp2plot}), suggesting robustness to training on code-mixed data and on using non-English pairs.
These results are broadly similar to Experiment 1, except that, as we see in Figure~\ref{fig:lmacc-iters}, the performance of the code-mixed data decays at a very different rate for the code-mixed data. Therefore, we used 16 iterations for both models. 
Why the code-mixed data is more robust to intervention is potentially interesting, but exploring it is beyond the scope of this work.

Table~\ref{tab:exp2} depicts the proportions of cases in which the probabilities move in the expected direction, and the results are similar. We see that there is not much change when pushing in the same direction as the template and larger changes when pushing in the opposite direction.
As with Experiment 1, this likely represents a ceiling effect.

%% file: tables/example.tex
\begin{table}[t]
\footnotesize
    \centering
    \begin{tabular}{p{0.4\linewidth} | p{0.50\linewidth}}
        \toprule
        \textbf{Most likely tokens pre-intervention} & friend, house, dream, novel, room, bed, book \\  \midrule
        \textbf{Most likely tokens after pushing to Spanish} & coma, car, man, la, son, del, más \\ \midrule
        \textbf{Most likely tokens after pushing to English} & house, dream, room, friend, book, tree, memory \\
        \bottomrule
    \end{tabular}
    \caption{Example of the most likely tokens predicted for the masked token pre and post-intervention for the English language text ``One day while Cat was wandering about, he came to a \texttt{[MASK]}.''}
    \label{tab:example}
\end{table}

%% file: sections/06-conclusion.tex
Overall, our results show that, if we take a sentence in Language A, embed it in a multilingual model, and use AlterRep to systematically push a particular word in that sentence towards Language B, the probability of words in Language B will go up. 
If we push a word in Language A towards Language A, there is little change except that, as shown in Table~5, highly probable words increase in probability overall.
Importantly, the probability of words in random control languages do not increase under either intervention.

What can we conclude from this?
First, since learning a language ID classifier can be used to causally affect the language of probable masked tokens, we take it as additional evidence \citep{libovicky-etal-2020-language,gonen-etal-2020-greek} that mBERT and XLM-R (and likely other models of similar structure) 
have both a language-specific and language-general component.
Second, this language-specific component is linearly extractable and  can be used causally to affect the language generated.
That said, we did not find evidence that it can be used for translation specifically since translation-equivalent words do not show a boost relative to controls.

In addition to shedding light on multilingual models, we think the method here shows that the AlterRep method \cite{ravfogel-etal-2021-counterfactual} can be fruitfully applied in settings beyond the syntactic application for which it was originally used.
In future work, we could use this method to explore linguistic typology in  multilingual model space.

\section*{Limitations}

Techniques like INLP extract information that is linearly extractable. While we've shown that it is possible to extract and manipulate language information using such simple linear techniques, more complex methods like those proposed by \citet{ravfogel-etal-2022-adversarial} might be able to manipulate more non-linearly encoded properties.

We have shown that language ID information is extractable and can be used to manipulate embeddings, but we urge caution in concluding that this means it could be used to practical effect (e.g., in machine translation). 
We leave the translation of these results into practical applications for future work.

The AlterRep procedure, as can be seen in our results and in \citet{ravfogel-etal-2021-counterfactual}, is sensitive to parameters like $\alpha$ and the number of INLP iterations. Picking these parameters is tricky and we have done it in a manner that preserves information in the language model.
It is possible that a different set of settings not explored here could lead to different results.

The risks associated with this work are the risks associated with any work dealing with large language models, including potential environmental impacts.

\section*{Acknowledgements}
 We thank Hila Gonen for helpful comments. 
 We thank Tal Linzen for presenting AlterRep to K.M.'s LIN 393 grad seminar and the students of that seminar for helpful comments.
 K.M. acknowledges funding from NSF Grant 2104995.

%% file: sections/99-appendix.tex
\section{Implementation}\label{app:implementation}

We use \texttt{bert-base-multilingual-cased} and \texttt{xlm-roberta-base} models from the Huggingface models  repository, and the \texttt{transformers} package for all of our probing experiments. Language ID classifiers were trained using \texttt{LinearSVC} classifier from \texttt{sklearn}. For training these classifiers, equal number of tokens from both labels were sampled. We used a batch size of 32, and a maximum sequence length of 256 when performing the intervention experiments.

\begin{figure*}
     \begin{subfigure}[b]{1\columnwidth}

    \centering
    \includegraphics[width=1\columnwidth]{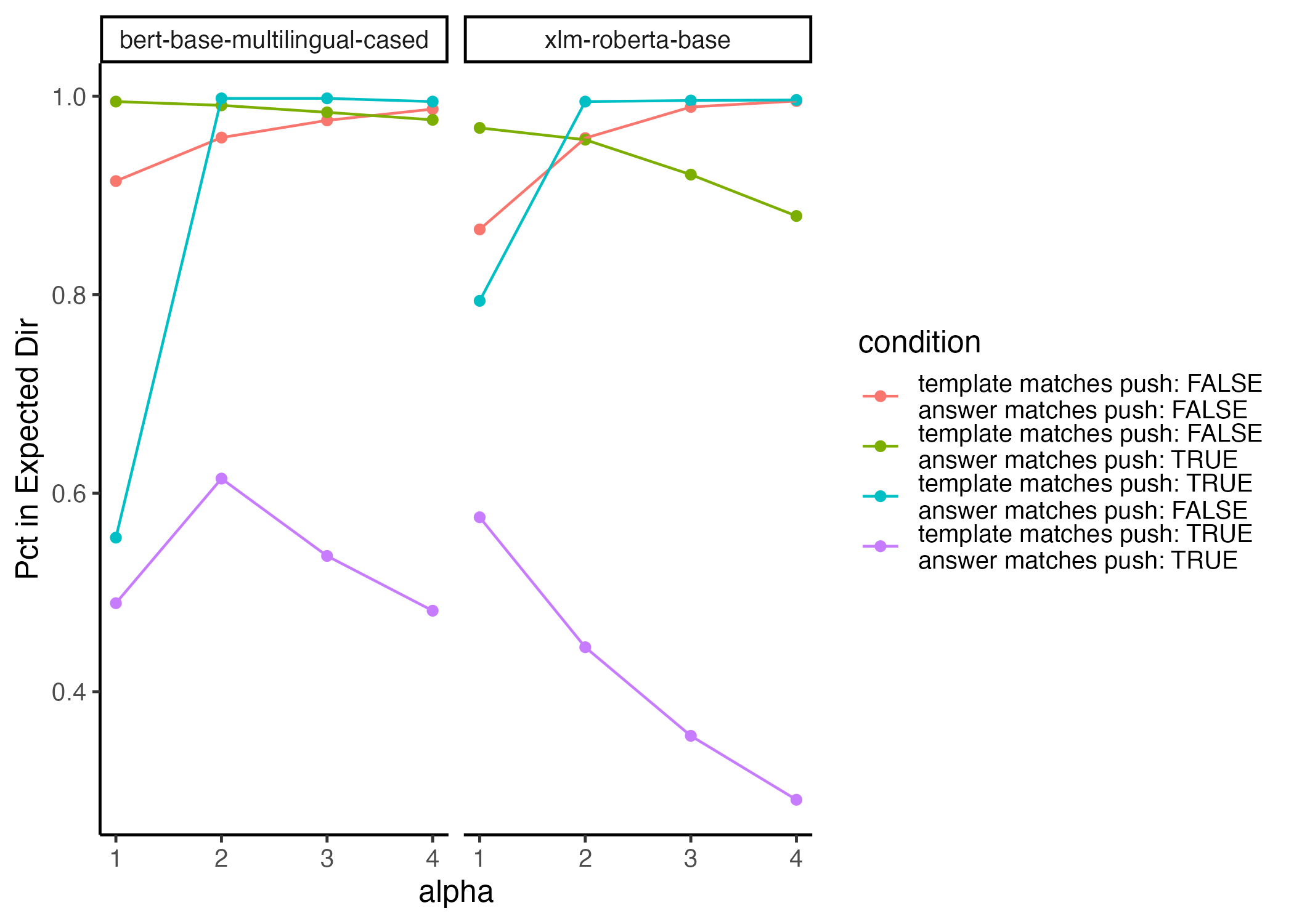}
        \end{subfigure}
     \begin{subfigure}[b]{1\columnwidth}
    \centering
    \includegraphics[width=1\columnwidth]{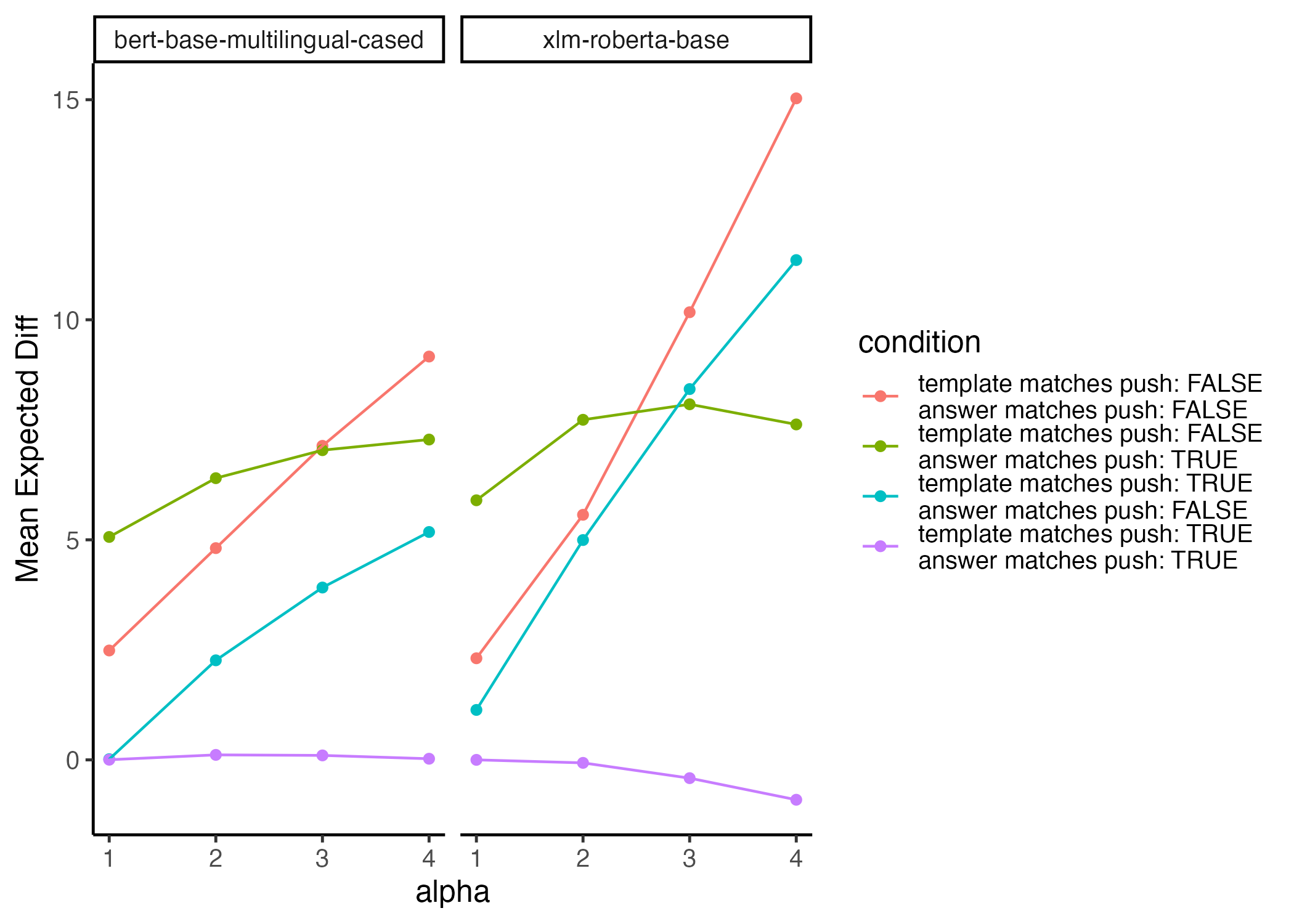}
        \end{subfigure}
        \caption{Left: Mean difference in log probability, across languages, in the \textit{expected} direction (positive if pushed to, negative if pushed away from) between before-intervention and after-intervention probabilities of either the pushed-to language or the pushed-away-from language, as a function of $\alpha$. Right: Proportion of the time, across languages, the intervention causes the probabilities to move in the \textit{expected} direction (positive if pushed \textit{to}, negative if pushed \textit{away from}), as a function of $\alpha$.}\label{fig:alpha}

\end{figure*}

\section{Effect of $\alpha$}\label{app:alpha}

For our Experiment 1 results, we plot key measures in Figure~\ref{fig:alpha} as a function of $\alpha$.
Specifically, we plot the proportion of the time we see movement in the expected direction and the mean change in log probability.

When $\alpha$ gets large, the words that we are pushing \textit{away from} continue to move in the expected direction.
This is likely because the increased shift can decrease the probability of those words arbitrarily, even while affecting the language model.

For words from the language that we are pushing \textit{towards}, there are diminishing returns to increasing $\alpha$ and in some cases we see decreases (as with the XLM-R purple line, which shows the probability of the target answer when we push towards its language).
This is likely because the target answer starts off with high probability, and larger $\alpha$ increasingly degrades the language model, causing the true answer to decrease in probability.